\documentclass[11pt]{article}

\usepackage[preprint]{acl}

\usepackage{times}
\usepackage{latexsym}

\usepackage[T1]{fontenc}

\usepackage[utf8]{inputenc}

\usepackage{microtype}

\usepackage{inconsolata}

\usepackage{graphicx}

\usepackage[dvipsnames]{xcolor}
\definecolor{cudOrange}{HTML}{D55E00}
\definecolor{cudBlue}{HTML}{005A9C}
\definecolor{cudGray}{HTML}{999999}
\definecolor{cudBrightOrange}{HTML}{E69F00}
\definecolor{cudSkyBlue}{HTML}{56B4E9}
\usepackage{amsmath,amssymb,mathtools}
\usepackage{tikz-dependency}
\usepackage{linguex}
\usepackage{xurl}
\usepackage{xspace}
\usepackage{subcaption}
\usepackage{booktabs}
\usepackage[capitalize,noabbrev]{cleveref}
\usepackage{cancel}
\usepackage{enumitem}
\usepackage{fontawesome5}

\newcommand{\wseq}[2]{\ensuremath{\boldsymbol w_{[#1: #2]}}\xspace}
\newcommand{\Wseq}[2]{\ensuremath{\boldsymbol W_{[#1: #2]}}\xspace}
\newcommand{\curid}{\ensuremath{k}\xspace}
\newcommand{\sstartid}{\ensuremath{1}\xspace}
\newcommand{\sendid}{\ensuremath{N}\xspace}
\newcommand{\nextid}{\ensuremath{\curid+1}\xspace}
\newcommand{\targetid}{\ensuremath{i}\xspace}

\newcommand{\wcur}{\ensuremath{w_{\curid}}\xspace}
\newcommand{\Wcur}{\ensuremath{W_{\curid}}\xspace}
\newcommand{\wtarget}{\ensuremath{w_{\targetid}}\xspace}

\newcommand{\context}{\ensuremath{\boldsymbol w_{[\sstartid:\curid)}}\xspace}
\newcommand{\Context}{\ensuremath{\boldsymbol W_{[\sstartid:\curid)}}\xspace}
\newcommand{\contextwotarget}{\ensuremath{\boldsymbol w_{[1:\curid)\setminus \targetid}}\xspace}

\newcommand{\sent}{\ensuremath{\wseq{\sstartid}{\sendid}}\xspace}
\newcommand{\future}{\ensuremath{\wseq{\curid}{\sendid}}\xspace}

\newcommand{\Futurenext}{\ensuremath{\Wseq{\nextid}{\sendid}}\xspace}

\newcommand{\Future}{\ensuremath{\Wseq{\curid}{\sendid}}\xspace}

\newcommand{\contextnext}{\ensuremath{\boldsymbol{w}_{[\sstartid:\curid+1)}}\xspace}

\newcommand{\contextwotargetnext}{\ensuremath{\boldsymbol{w}_{[\sstartid:\curid+1)\setminus\targetid}}\xspace}
\newcommand{\FutureNext}{\ensuremath{\Wseq{\nextid}{\sendid}}\xspace}

\newcommand{\past}{\ensuremath{\boldsymbol w_{[\sstartid:\targetid)}}\xspace}
\newcommand{\intervenings}{\ensuremath{\boldsymbol w_{(\targetid:\curid)}}\xspace}

\newcommand{\mask}{\texttt{[M]}\xspace}

\newcommand{\logtwo}{\ensuremath{\log_{2}}\xspace}
\newcommand{\pmi}{\ensuremath{\operatorname{pmi}}\xspace}
\newcommand{\CHMI}{\ensuremath{\mathcal{P}_{\mathrm{pred}}}\xspace}
\newcommand{\chmi}[2]{\ensuremath{\CHMI(#1;#2)}\xspace}
\newcommand{\stor}[1]{\ensuremath{\operatorname{InfoStor}_{#1}}\xspace}

\newcommand{\DLT}{\textcolor{cudOrange}{\textsc{DLT}}\xspace}
\newcommand{\InfoDLT}{\textcolor{cudBrightOrange}{\textsc{DLT}-\textsc{on}-\textsc{Info}}\xspace}
\newcommand{\Info}{\textcolor{cudBlue}{\textsc{Info}}\xspace}
\newcommand{\DLTInfo}{\textcolor{cudSkyBlue}{\textsc{Info}-\textsc{on}-\textsc{DLT}}\xspace}

\newcommand{\dll}{\ensuremath{\Delta_{\mathit{LL}}}\xspace}

\newcommand{\prob}[1]{\ensuremath{p(#1)}\xspace}
\newcommand{\condprob}[2]{\ensuremath{p(#1\mid #2)}\xspace}

\newcommand{\condpmi}[3]{\ensuremath{\pmi(#1;#2\mid #3)}\xspace}
\newcommand{\mi}[2]{\ensuremath{\operatorname{I}(#1;#2)}\xspace}

\newcommand{\condE}[3]{\ensuremath{\underset{#1\sim\condprob{\cdot}{#2}}{\mathbb{E}}\!\big[#3\big]}\xspace}
\newcommand{\E}[2]{\ensuremath{\mathbb{E}_{#1}\Big[#2\Big]}\xspace}
\newcommand{\KL}[2]{\ensuremath{D_{\operatorname{KL}}\Big(#1\parallel #2\Big)}\xspace}

\title{Information-Theoretic Storage Cost in Sentence Comprehension}

\author{
    Kohei Kajikawa${}^1$\quad Shinnosuke Isono${}^2$\quad Ethan Gotlieb Wilcox${}^1$ \\
    ${}^1$Department of Linguistics, Georgetown University, USA \\
    ${}^2$National Institute for Japanese Language and Linguistics, Japan \\
    \small{
        \textbf{Correspondence:} \href{mailto:kk1571@georgetown.edu}{\texttt{kk1571@georgetown.edu}}
    }
  }

\begin{document}

\maketitle

\begin{abstract}

Real-time sentence comprehension imposes a significant load on working memory, as comprehenders must maintain contextual information to anticipate future input. While measures of such load have played an important role in psycholinguistic theories, they have largely been formalized using symbolic grammars, which assign discrete, uniform costs to syntactic predictions. This study proposes a measure of processing storage cost based on an information-theoretic formalization, as the amount of information previous words carry about future context, under uncertainty. Unlike previous discrete, grammar-based metrics, this measure is continuous, probabilistic, theory-neutral, and can be estimated from pre-trained neural language models. The validity of this approach is demonstrated through three analyses in English: our measure (i) recovers well-known processing asymmetries in center embeddings and relative clauses, (ii) correlates with a grammar-based storage cost in a syntactically-annotated corpus, and (iii) predicts reading-time variance in two large-scale naturalistic datasets over and above baseline models with traditional information-based predictors.
Our code is available at \faGithub~\url{https://github.com/kohei-kaji/info-storage}.

\end{abstract}

\section{Introduction}
\label{sec:intro}

A large body of evidence shows that language comprehension is highly incremental~\citep{marslen-wilson-1973,tanenhaus-etal-1995,kamide-etal-2003}.
As the input unfolds, the parser integrates incoming words into the evolving representation while generating expectations about upcoming content.
However, since working memory capacity is limited, maintaining these expectations incurs a cognitive cost:
as unresolved predictions accumulate, they consume memory resources and create a processing bottleneck~\citep{just-carpenter-1992}.
It has been argued that this bottleneck gives rise to processing difficulty, for example, in center-embedding sentences such as \ref{ex:center-embed}, below, where the processing of an outer clause is interrupted by an inner one, forcing the parser to hold the incomplete outer structure in working memory~\citep{yngve-1960,miller-chomsky-1963,gibson-1998}:
\ex. The reporter \texttt{[}who the senator \texttt{[}who Mary met\texttt{]} attacked\texttt{]} ignored the president. \label{ex:center-embed}

The memory burden of maintaining predicted syntactic elements has been discussed previously, often under the more general term of \emph{storage cost}~\citep{yngve-1960,miller-chomsky-1963,abney-johnson-1991,rambow-joshi-1994,gibson-1998,gibson-2000,kobele-etal-2013,isono-2024}.
The hypothesis that storage cost actively influences online processing has received empirical support in both controlled experiments across languages~\citep{chen-etal-2005,nakatani-gibson-2010,stepanov-stateva-2015,ristic-etal-2022} and naturalistic reading~\citep{isono-etal-ur,isono-kajikawa-2026}.

\begin{figure}[t]
    \centering
    \includegraphics[width=\linewidth]{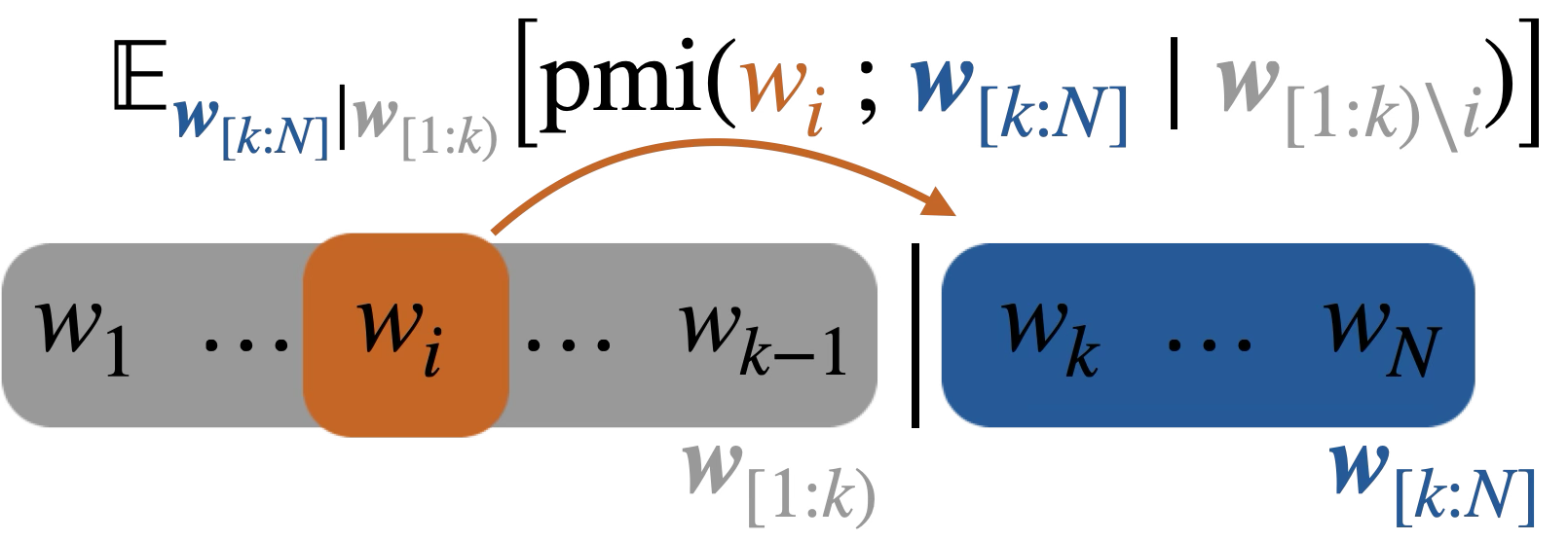}
    \caption{Illustration of the proposed storage cost measure.
    It quantifies the predictive potential shared between a \textcolor{cudOrange}{target word \wtarget} and the \textcolor{cudBlue}{future \future}, conditioned on the \textcolor{cudGray}{remaining context}.
    Information-theoretic storage cost at \wcur is defined as the sum of the predictive potentials across all context words, representing the load of pending information.}
    \label{fig:pmi_illustration}
\end{figure}

\begin{figure*}[t]
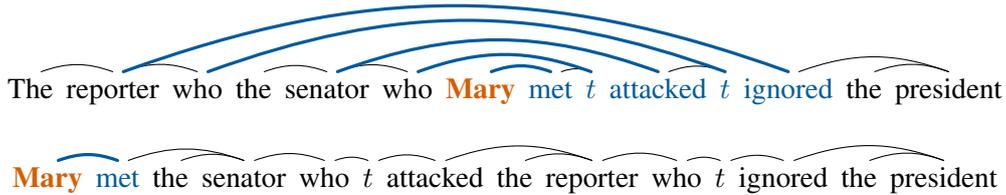

    \centering
    \begin{dependency}[theme=simple, edge style={-}, arc angle=20]
        \begin{deptext}
            The \& reporter \& who \& the \& senator \& who \& \textcolor{cudOrange}{\textbf{Mary}} \& \textcolor{cudBlue}{met} \& \textcolor{cudBlue}{$t$} \& \textcolor{cudBlue}{attacked} \& \textcolor{cudBlue}{$t$} \& \textcolor{cudBlue}{ignored} \& the \& president \\
        \end{deptext}
        \depedge{6}{9}{}
        \depedge{3}{11}{}
        \depedge{2}{12}{}
        \depedge{5}{10}{}
        \depedge{7}{8}{}
        \depedge{2}{3}{}
        \depedge{5}{6}{}
        \depedge{1}{2}{}
        \depedge{4}{5}{}
        \depedge{13}{14}{}
        \depedge{12}{14}{}
        \depedge{8}{9}{}
        \depedge{10}{11}{}
        \depedge[edge style={very thick, cudBlue}]{2}{12}{}
        \depedge[edge style={very thick, cudBlue}]{3}{11}{}
        \depedge[edge style={very thick, cudBlue}]{6}{9}{}
        \depedge[edge style={very thick, cudBlue}]{5}{10}{}
        \depedge[edge style={very thick, cudBlue}]{7}{8}{}
    \end{dependency}

    \begin{dependency}[theme=simple, edge style={-}, arc angle=20]
        \begin{deptext}
            \textcolor{cudOrange}{\textbf{Mary}} \& \textcolor{cudBlue}{met} \& the \& senator \& who \& $t$ \& attacked \& the \& reporter \& who \& $t$ \& ignored \& the \& president \\
        \end{deptext}
        \depedge{1}{2}{}
        \depedge{2}{4}{}
        \depedge{4}{5}{}
        \depedge{3}{4}{}
        \depedge{5}{6}{}
        \depedge{6}{7}{}
        \depedge{7}{9}{}
        \depedge{8}{9}{}
        \depedge{9}{10}{}
        \depedge{10}{11}{}
        \depedge{11}{12}{}
        \depedge{12}{14}{}
        \depedge{13}{14}{}
        \depedge[edge style={very thick, cudBlue}]{1}{2}{}
    \end{dependency}

    \caption{Illustration of DLT storage cost based on the predicted syntactic head hypothesis.
    In the center-embedded structure (top), the storage cost at \textcolor{cudOrange}{\textit{Mary}} is \textbf{five} memory units, because five syntactic heads (the \textcolor{cudBlue}{blue}-colored words) are predicted to form a grammatical sentence at \textcolor{cudOrange}{\textit{Mary}}.
    In contrast, the right-branching structure (bottom) imposes a minimal storage load.
    $t$ denotes a trace of extraction.
    This example is adopted from \citet{chen-etal-2005}.}
    \label{fig:pred_heads}
\end{figure*}

Previously, most formalizations of storage costs have been derived from the incremental states of symbolic parsers \citep[e.g.,][]{gibson-2000}.
While grammar-based storage cost successfully captures the memory burden of structural prediction, it relies on specific syntactic theories and assumes discrete, uniform costs for predicted elements.
In this paper, we propose a grammar-independent reformulation of storage cost grounded in information theory.
The key idea is that maintaining a syntactic prediction corresponds to retaining information about context words that are relevant to future input.
We argue that this measure can be formalized as the extent to which the context words reduce the surprisal of future material under uncertainty.
Taken in context, this is therefore the \emph{contextualized half-pointwise mutual information} between words in context and future words.
The resulting measure requires no explicit syntactic formalism, can be estimated from neural language models, and is continuous, probabilistic, and quantified in bits rather than discrete counts.

We validate the proposal through three complementary analyses.
First, we examine whether the measure recovers well-known processing asymmetries in center-embedding and relative clause sentences, establishing theoretical plausibility.
Second, we confirm a positive correlation with grammar-based storage cost, showing that the two measures capture overlapping structural intuitions.
Third, we evaluate its predictive power on two large-scale naturalistic English reading-time datasets and compare it to that of a grammar-derived measure.
We find significant improvements across multiple reading-time measures, establishing information-theoretic storage cost as a viable, theory-neutral alternative.
At the same time, the variance each measure explains is largely independent, suggesting that storage cost estimated from formal grammars and neural language models accounts for partially distinct sources of reading-time variance.

\section{Grammar-based Storage Cost}
\label{sec:background}

The idea that maintaining predicted syntactic elements consumes working memory resources has long been assumed in psycholinguistics across different grammar formalisms~\citep{yngve-1960,miller-chomsky-1963,abney-johnson-1991,rambow-joshi-1994,gibson-1998,gibson-2000,kobele-etal-2013,isono-2024}.
As an example, we take Dependency Locality Theory~\citep[DLT;][]{gibson-2000}.
Grounded in dependency grammar, DLT defines storage cost as the number of predicted syntactic heads required to complete the current input as a grammatical sentence.

Consider the following center-embedded sentence~\ref{ex:ce} and its right-branching variant~\ref{ex:rb}:%
\ex.\label{ex:cerb}
    \a. The reporter \texttt{[}who the senator \texttt{[}who \textcolor{cudOrange}{\textbf{Mary}} met\texttt{]} attacked\texttt{]} ignored the president.\label{ex:ce}
    \b. Mary met the senator \texttt{[}who attacked the reporter \texttt{[}who ignored the president\texttt{]}\texttt{]}.\label{ex:rb}

At the point of \textcolor{cudOrange}{\textit{Mary}} in~\ref{ex:ce}, five syntactic heads are required to form a grammatical sentence: (i) three verbs to which the three subject NPs connect, and (ii) two extraction traces anticipated in object positions (see also \cref{fig:pred_heads}).
Thus, DLT storage cost at \textcolor{cudOrange}{\textit{Mary}} incurs a cost of five ``memory units.''
In contrast, at each word in the right-branching variant of this sentence~\ref{ex:rb}, at most one syntactic head is unresolved.
Consequently, DLT storage cost correctly predicts the well-known difficulty of processing center-embedded structures compared to right-branching ones~\citep{yngve-1960,miller-chomsky-1963,gibson-1998}.

\paragraph{Storage cost vs.\ integration cost.}
Storage cost in DLT constitutes a complementary component to \emph{integration cost}.
While storage cost quantifies the memory burden of maintaining structural predictions before they are resolved, integration cost arises from the resource consumption required to resolve syntactic dependencies when new input is encountered.
They represent different cognitive demands imposed by the same dependency structure.
Integration cost has long been a focal point of research~\citep[e.g.,][]{gibson-2000,futrell-etal-2020-dependency} because it offers a straightforward account of \emph{locality effects}, the processing difficulty associated with longer syntactic distances~\citep{gibson-2000,grodner-gibson-2005,staub-2010,bartek-etal-2011,roland-etal-2021}.
However, its empirical robustness is debated because null effects or even unexpected negative correlations (i.e., facilitation) are reported in naturalistic reading~\citep{demberg-keller-2008,shain-schuler-2018,dotlacil-2021,isono-2024}.
In this context, \emph{lossy-context surprisal}~\citep{futrell-etal-2020-lossy}, defined as prediction error resulting from noisy context representations, can be seen as an information-theoretic generalization of integration cost. 
Within this framework, locality effects emerge as a consequence of memory noise.

\paragraph{Linking storage cost to reading behavior.}
The linking hypothesis between storage cost and reading times is that storage and integration share the same pool of cognitive resources: as more resources are allocated to storage, fewer remain available for integration, resulting in slower reading times~\citep{gibson-1998,gibson-2000}.
Controlled experiments have validated this prediction across languages, including English, Japanese, Slovenian, and Spanish~\citep{chen-etal-2005,nakatani-gibson-2010,stepanov-stateva-2015,ristic-etal-2022}.
Furthermore, \citet{isono-etal-ur} demonstrate that storage costs based on both dependency grammar\footnote{\citet{isono-etal-ur} adopt the Universal Dependencies framework~\citep{de-marneffe-etal-2021} and operationalize DLT storage cost as the number of unseen tokens whose co-dependents are already seen at a given word.} and combinatory categorial grammar~\citep[CCG;][]{steedman-2000} have larger predictive power on a held-out test set of naturalistic Japanese reading times than integration cost.

\paragraph{Limitations of grammar-based storage cost.}
Storage cost metrics based on symbolic grammars have several limitations.
Theoretically, they assume that all syntactic predictions carry the same cost.
Practically, it can be challenging to count the heads needed to complete a sentence.
This often requires specific assumptions about grammatical completion and accurate representation of phonologically null elements (e.g., traces), which are absent in surface-based annotations like Universal Dependencies~\citep{de-marneffe-etal-2021}.
In addition, grammar-based metrics are often insensitive to temporal structural ambiguity, as they evaluate memory load based on a single, fully resolved parse tree.
Our information-theoretic approach eliminates the need for specific syntactic theories and assumptions while capturing the probabilistic nature of sentence comprehension and replacing the discrete fixed-cost assumption with a continuous measure quantified in bits.

\section{Information-Theoretic Storage Cost}
\label{sec:proposal}

This section develops an information-theoretic formalization of storage cost.
We first motivate the approach and provide the formal definition, then describe how to estimate the measure using a pre-trained masked language model.

\subsection{From Syntactic to Information Storage}

Grammar-based storage cost counts the number of predicted syntactic elements that must be held in memory.
The implicit assumption is that each prediction represents information about the future that the comprehender must maintain in working memory until it is resolved and integrated into a longer-term memory store.
We propose that this can be directly quantified as the amount of information content that a word carries about future words, or the extent to which that word \emph{reduces} the surprisal of future words.
We view information-based storage as a generalization of grammar-based storage cost.
Grammar-based approaches are a special case where (i) information about the future is mediated exclusively by syntactic predictions and (ii) each prediction carries a uniform cost.
Our formalization relaxes both assumptions---information is measured over surface forms, and each word's contribution is a continuous quantity in bits.

\subsection{Predictive Potentials as Expected PMI}
\label{sec:predinfo}
We operationalize the information a word carries about the future using \emph{contextualized pointwise mutual information} \citep[\pmi;][]{hoover-etal-2021-linguistic}.
The \pmi measures the association between outcomes of random variables (RVs) $X$ and $Y$ as the log-ratio of their joint probability to the product of their marginals:
\begin{align}
    \pmi(x;y) \coloneqq \logtwo\frac{p(X=x,Y=y)}{p(X=x)p(Y=y)}
\end{align}

Consider a word-valued RV $W$, which ranges over a vocabulary $\Sigma$, and a sentence-valued RV $\boldsymbol{W}$, which ranges over strings drawn from its Kleene closure, $\Sigma^*$.
For a sentence $\sent = [w_1, w_2, \ldots, w_N$], at position \curid, the observed context is $\context = [w_1, \ldots, w_{k-1}]$, and the future sequence is $\future = [w_k, \ldots, w_N$].
For a target word at position \targetid (where $\targetid < \curid$), we write the context excluding \wtarget as $\contextwotarget = [w_1, \ldots, w_{i-1}, w_{i+1}, \ldots, w_{k-1}]$.

The contextualized PMI between \wtarget and the future sequence \future, given the rest of the context \contextwotarget, is defined as:
\begin{align}
    & \condpmi{\wtarget}{\future}{\contextwotarget} \\
    &\coloneqq \logtwo \frac{\condprob{\wtarget, \future}{\contextwotarget}}{\condprob{\wtarget}{\contextwotarget} \condprob{\future}{\contextwotarget}} \nonumber \\
    & = \logtwo \frac{\condprob{\future}{\context}}{\condprob{\future}{\contextwotarget}}. \nonumber
\end{align}

This quantity measures how much observing \wtarget changes our prediction of the future \future, in bits.

Since the future sequence \future is unknown during left-to-right processing, we consider the expectation over possible continuations.
We term this expected value \emph{predictive potential} denoted as \CHMI:
\begin{align}
    & \chmi{\wtarget}{\Future} \label{eq:predpotential}\\
    &\coloneqq \condE{\future}{\context}{\condpmi{\wtarget}{\future}{\contextwotarget}}, \nonumber
\end{align}
where the expectation is taken with respect to \Future conditioned on \context.
This quantity is formally the \emph{contextualized half-pointwise mutual information} and is equivalent to the Kullback-Leibler (KL) divergence between the predictive distributions with and without \wtarget (see~\cref{app:kl}).
\Cref{fig:pmi_illustration} provides a visual illustration of this measure.

\paragraph{Monotonic decay.}
The predictive potential of \wtarget regarding the future sequence is updated as processing proceeds.
As the distance to the future increases, this information quantity is monotonically non-increasing in expectation (see~\cref{app:mono} in detail).
This property aligns with the activation decay of memory traces over time~\citep{lewis-vasishth-2005,lewis-etal-2006}.

\paragraph{Information storage.}
We define the \emph{information storage} at position \curid, representing the total memory load, as the sum of the predictive potentials from all preceding words:
\begin{align} \label{eq:info_storage}
    \stor{\curid} \coloneqq \sum_{\targetid=\sstartid}^{\curid-1} \chmi{\wtarget}{\Future}.
\end{align}

It is this total information storage term that we predict is causally connected to reading difficulty.

\subsection{Estimation using Masked LLMs}
\label{sec:bert}

We estimate \CHMI using BERT \citep{devlin-etal-2019},  a bidirectional masked language model.
For each pair $(\targetid, \curid)$ with $\targetid < \curid$, we construct inputs with and without \wtarget masked:
\begin{align*}
    & \text{with \wtarget:} \quad \past,\; \wtarget,\; \intervenings,\; \underbrace{\mask, \ldots, \mask}_{\future} \\
    & \text{without \wtarget:} \quad \past,\; \mask,\; \intervenings,\; \underbrace{\mask, \ldots, \mask}_{\future}
\end{align*}
where \mask denotes a mask token and each word in \future is replaced by the appropriate number of mask tokens.\footnote{We add \texttt{[CLS]} at the beginning and \texttt{[SEP]} at the end.}

BERT provides token-level distributions but not joint distributions over multiple tokens.
We assume conditional independence among masked positions.
Let $\boldsymbol{m}$ denote the vector of masked token positions in the future region, and let $q^{\wtarget+}_m$ and $q^{\wtarget-}_m$ be BERT's predictive distributions at position $m$ with and without \wtarget visible, respectively.
The KL divergence then decomposes as:
\begin{align}
    & \chmi{\wtarget}{\Future} \approx \sum_{m \in \boldsymbol{m}} \KL{q^{\wtarget+}_m}{q^{\wtarget-}_m}. 
\end{align}

We use \texttt{bert-base-uncased} from the Transformers library~\citep{wolf-etal-2020-transformers} in this study, in line with prior work that estimates contextualized \pmi using BERT~\citep{hoover-etal-2021-linguistic,wilcox2024regressions}.\footnote{A supplementary analysis using BERT-Large (\texttt{bert-large-uncased}) and RoBERTa (\texttt{roberta-base};~\citealp{liu2019roberta}) is provided in \cref{app:bertlargeroberta}.
Note that the overall findings remain largely robust to the choice of model. Specifically, RoBERTa yields highly comparable results to the BERT-base model in both the correlation analysis with DLT and reading-time modeling. BERT-Large also exhibits broadly similar trends, with only minor variations.
}
We note that the assumption of conditional independence is strong (see \cref{sec:limitation} for future directions).
We view the BERT-based methods outlined above, therefore, as a starting point which can be improved upon in future work.

\section{Experiments and Results}

We evaluate the information storage measure presented in Equation \eqref{eq:info_storage} through three analyses: (i) we visualize its distribution on specific constructions; (ii) we inspect its correlation with DLT-based storage measures; and (iii) we test its predictive power for human naturalistic reading data.

\subsection{Case Studies: Center Embeddings and Relative Clauses}
\label{sec:illustration}

To evaluate the behavior of the proposed measure, we visualize estimated storage costs for two classic syntactic asymmetries much discussed in the psycholinguistics literature.
These are center-embedding vs.\ right-branching structures and subject vs.\ object relative clauses.\footnote{
    Note that while DLT integration cost also accounts for these asymmetries, it predicts a different locus of processing difficulty compared to storage cost. For instance, in center-embedding structures, DLT storage cost peaks at the most deeply embedded noun phrase, whereas DLT integration cost peaks at the the main verb.
}
For each case, we procedurally generate sets of stimulus sentences and plot the mean information storage at each word position across these items.
The full list of generated materials is provided in~\cref{app:sents}.

\paragraph{Center embedding.} 
Center-embedded structures (CE) incur notorious processing difficulty compared to right-branching (RB) variants~\citep{yngve-1960,miller-chomsky-1963,gibson-1998}. 
We generate 30 sentence pairs (items) using fixed templates with placeholders for animate nouns ($N$) and past-tense transitive verbs ($V$) requiring animate arguments:
\begin{description}[noitemsep]
    \item[CE] \textit{The} $N_1$ \textit{who the} $N_2$ \textit{who the} $N_3$ $V_3$ $V_2$ $V_1$ \textit{the} $N_4$
    \item[RB] \textit{The} $N_3$ $V_3$ \textit{the} $N_2$ \textit{who} $V_2$ \textit{the} $N_1$ \textit{who} $V_1$ \textit{the} $N_4$
\end{description}

\begin{figure}[t]
    \centering
    \begin{subfigure}[b]{\linewidth}
        \centering
        \includegraphics[width=\linewidth]{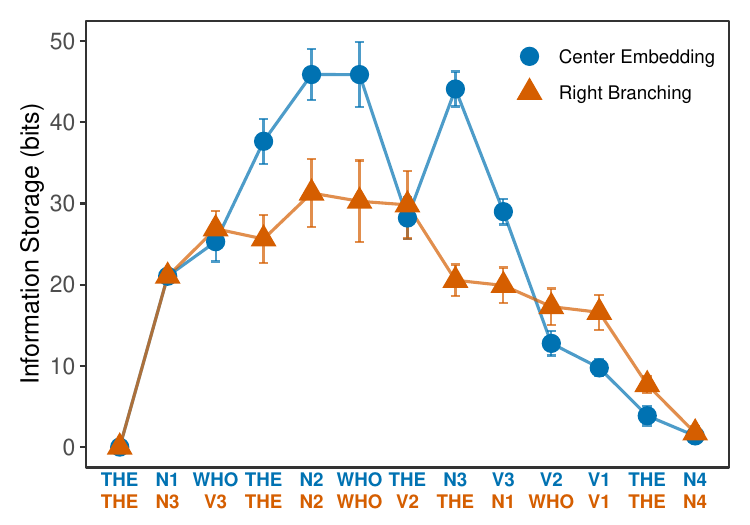}
        \caption{Center embedding (blue) vs.\ right branching (orange).}
        \label{fig:embedding_storage}
    \end{subfigure}
    \vspace{0.4cm}
    \begin{subfigure}[b]{\linewidth}
        \centering
        \includegraphics[width=\linewidth]{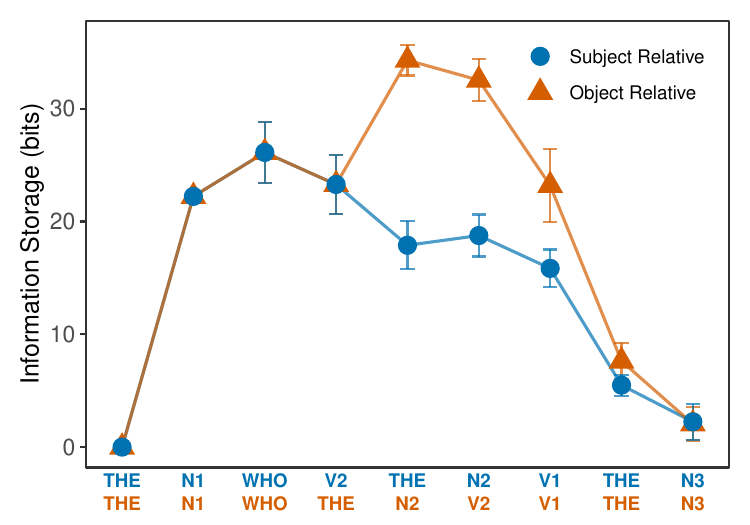}
        \caption{Subject relative (blue) vs.\ object relative (orange).}
        \label{fig:rc_storage}
    \end{subfigure}
    \caption{Mean information storage estimated by BERT at each word position. Error bars represent 95\% confidence intervals across 30 items. In both cases, the more difficult structure (CE, ORC) exhibits higher storage cost, consistent with behavioral asymmetries.}
    \label{fig:combined_analysis}
\end{figure}

As shown in \cref{fig:embedding_storage}, information storage rises sharply in CE structures as unresolved dependencies accumulate, whereas RB structures remain lower. Summed across all words, the total information storage for CE ($\mu=303.43$ bits, $\sigma=44.42$) was substantially higher than for RB ($\mu=250.54$ bits, $\sigma=48.54$).

\paragraph{Subject/object relatives.}
Object relative clauses (ORC) are consistently more difficult to process than subject relative clauses (SRC), as evidenced by longer reading times across multiple experimental paradigms \citep{king-just-1991,grodner-gibson-2005,staub-2010,vani-etal-2021}.
We generate 30 items based on the following templates:

\begin{description}[noitemsep]
    \item[SRC] \textit{The} $N_1$ \textit{who} $V_2$ \textit{the} $N_2$ $V_1$ \textit{the} $N_3$
    \item[ORC] \textit{The} $N_1$ \textit{who the} $N_2$ $V_2$ $V_1$ \textit{the} $N_3$
\end{description}

Results are visualized in \Cref{fig:rc_storage}.
Our formulation predicts a critical difference between the conditions emerging at the embedded noun phrase.
In ORC, the predictive potential of preceding words (e.g., \textit{who}, \textit{the}) accumulates, causing a peak nearly twice as high as the maximum in SRC. Consequently, ORCs yielded higher total information storage ($\mu=171.35$ bits, $\sigma=21.12$) than SRCs ($\mu=131.87$ bits, $\sigma=20.70$).

In both cases, our information-theoretic measure recovers the well-documented processing asymmetries from distributional statistics alone, without relying on syntactic annotation.

\subsection{Correlation with DLT}
\label{sec:correlation}

As a second validation of our proposal, we examine whether information storage correlates with the grammar-based DLT storage cost. 
We predict that correlations should be positive and moderate, as information-based storage cost contains semantic and other non-structural information not captured by purely grammar-based metrics.

We use the UD\_English-GUM corpus~\citep{Zeldes2017}, a manually annotated treebank comprising 13,263 sentences and 233,926 words.
We compute both DLT storage cost and information storage for each sentence.
The original definition of DLT storage cost is the number of predicted syntactic heads required to complete the current input as a grammatical sentence.
Since it is challenging to apply this definition to large-scale corpora (as discussed in~\cref{sec:background}), we operationalize it as the count of unseen tokens whose co-dependents have been encountered.
In this calculation, we exclude the following dependency relations: \texttt{punct}, \texttt{root}, \texttt{dep}, and \texttt{reparandum}.
For information storage, to align BERT's subword tokenization with UD token boundaries, we define words by whitespace and sum token-level values within each word.

\begin{figure}[t]
    \centering
    \includegraphics[width=\linewidth]{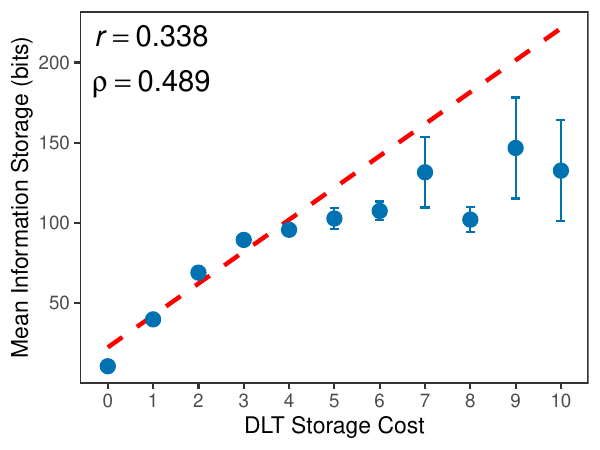}
    \caption{Mean information-theoretic storage cost as a function of DLT storage cost in the UD\_English-GUM corpus~\citep{Zeldes2017}. Points represent the mean value for each DLT bin with 95\% confidence intervals. For this visualization, bins with fewer than 100 observations were excluded due to data sparsity. The red dashed line represents the linear regression fitted to the raw data corresponding to the displayed bins.}
    \label{fig:dlt_correlation}
\end{figure}

\Cref{fig:dlt_correlation} visualizes the relationship between the two measures.
We observe a moderate positive correlation (Pearson's $r = 0.34$, Spearman's $\rho = 0.49$). Note, however, that the means of the DLT bins suggest that the underlying relationship may be sub-linear.
This sub-linearity likely reflects a fundamental difference between the two metrics: while DLT storage increases linearly as it counts the number of unseen tokens required by the co-dependents in context based on gold parses, information storage may not do so because mutual information in natural language decays according to a power law over distance~\citep[e.g.,][]{debowski-2015,hahn-etal-2021}.
Regardless, the correlations suggest that our information storage captures the core intuition behind grammar-based storage cost using only distributional statistics.

\subsection{Naturalistic Reading Times}\label{sec:rtanalysis}
We evaluate whether information storage improves reading-time prediction in two of the largest naturalistic English reading-time datasets currently available ($N$$\approx$100--200 each): Natural Stories~\citep{futrell-etal-2021} and OneStop~\citep{berzak-etal-2025onestop}.
We also examine whether information storage explains reading-time variance above and beyond DLT storage by testing their respective contributions when the other measure is already included in the model.

\subsubsection{Data}

Natural Stories~\citep{futrell-etal-2021} consists of 10 naturalistic narratives and 10,256 words.
We use self-paced reading (SPR) times from 181 native English speakers~\citep{futrell-etal-2021} and A-Maze reading times\footnote{A-Maze is a variant of the Maze task \citep{forster2009maze}, in which, at each sentence position, participants must choose between a ``true'' next-word continuation and a distractor. Choice times are taken as a proxy for incremental processing times.} from 95 native English speakers~\citep{boyce-levy-2023}.
OneStop~\citep{berzak-etal-2025onestop} consists of 10 articles and 35,181 words with eye-tracking data from 180 native English speakers (the ``ordinary reading'' sub-portion is used).
Following standard practice, we examine three eye-tracking measures: first-pass duration (FPD), the sum of all fixations from first entering a region until the first exit in either direction; go-past duration (GPD), the sum of all fixations until the first exit to the right; and total fixation duration (TFD), the sum of all fixations in a region, including re-reading. 

We apply several preprocessing steps. For all datasets, we exclude the first and last words of each sentence and any words containing punctuation.
For Natural Stories, we apply different criteria for each task.
For self-paced reading, we follow \citet{futrell-etal-2021} by excluding reading times shorter than 100 ms or longer than 3,000 ms, as well as participants with low comprehension accuracy (fewer than 5$/$6 correct).
For the A-Maze task, participants with less than 80\% accuracy are excluded~\citep{boyce-levy-2023}.

We model the mean reading time across participants~\citep{smith-levy-2013,goodkind-bicknell-2018,wilcox-etal-2023-language,wilcox-etal-2023-testing}.
Following \citet{wilcox-etal-2023-language,wilcox-etal-2023-testing}, we treat skipped words in eye-tracking data as having zero ms.

\subsubsection{Statistical analysis}
Following standard statistical model comparison and prior work on naturalistic reading-time analysis~\citep[e.g.,][]{frank-bod-2011,goodkind-bicknell-2018,wilcox-etal-2023-testing}, we test whether adding storage cost measures to a baseline linear regression model improves reading-time prediction.
The baseline model includes word positions (in the sentence and document), word length (the number of characters), unigram surprisal, and GPT-2 surprisal.
DLT storage is calculated as the number of unseen tokens whose co-dependents are already seen at a given word.
To obtain these dependencies, we use UD parses in Natural Stories~\citep{futrell-etal-2021}, and parses generated by \texttt{Stanza}~\citep{qi-etal-2020} for OneStop.
Unigram surprisal is estimated using the implementation by \citet{oh-etal-2024} on approximately 33 billion pre-tokenized tokens from the Pile dataset~\citep{gao-etal-2020}.
For GPT-2 surprisal, we use the 124M-parameter GPT-2~\citep{radford-etal-2019}, as it is a better predictor of reading times than larger models~\citep{oh-schuler-2023,shain-etal-2024}.
To estimate GPT-2 surprisal, we use a context of up to 1,024 preceding tokens and adopt whitespace-trailing decoding~\citep{oh-schuler-2024}, which reassigns the probability of a leading whitespace to the preceding word.
We include spillover terms for word length, unigram surprisal, GPT-2 surprisal, and both storage measures.
All predictors are $z$-scored.
Note that the storage measures are not highly correlated with other predictors. The full correlation matrix is provided in~\cref{fig:large_cor} of~\cref{app:cor}.\footnote{
    As information storage exhibits high autocorrelation ($r = 0.85$ in both datasets), we compute the Variance Inflation Factor (VIF) for all predictors to ensure that multicollinearity does not destabilize the regression estimates.
    The VIF values peak at approximately 3.8 for information storage and its spillover term. This remains well below the conservative threshold of 5, which indicates that essentially no additional confounding collinearity exists.
}

To assess the predictive power of the storage measures for reading-time variance, we evaluate the change in predictive power under four conditions:
(i) adding information storage to the baseline (\Info),
(ii) adding DLT storage to the baseline (\DLT),
(iii) adding information storage to the baseline that already includes DLT storage (\DLTInfo), and
(iv) adding DLT storage to the baseline that already includes information storage (\InfoDLT).
Conditions \Info and \DLT test the independent contribution of each measure, while \DLTInfo and \InfoDLT test whether each measure provides additional predictive power when the other is controlled for.

We evaluate predictive power using the per-word change in log-likelihood (\dll) between the target and baseline models. 10-fold cross-validation (CV) is employed to estimate \dll on held-out test data. Significance is assessed via a one-sided permutation test (20,000 iterations) using the mean \dll as the test statistic. To account for multiple comparisons across all 20 tests (5 datasets $\times$ 4 conditions), the Benjamini-Hochberg procedure~\citep{benjamin-hochberg-1995} is applied to control the false discovery rate (FDR) at $\alpha=0.05$.
We also examine the mean coefficient for information storage and DLT storage (averaged across the 10 CV folds) in the \Info and \DLT conditions to verify whether they are positive, as hypothesized.

\subsubsection{Results}

\begin{figure*}[t]
    \centering
    \begin{subfigure}[t]{0.48\textwidth}
        \centering
        \includegraphics[width=\linewidth]{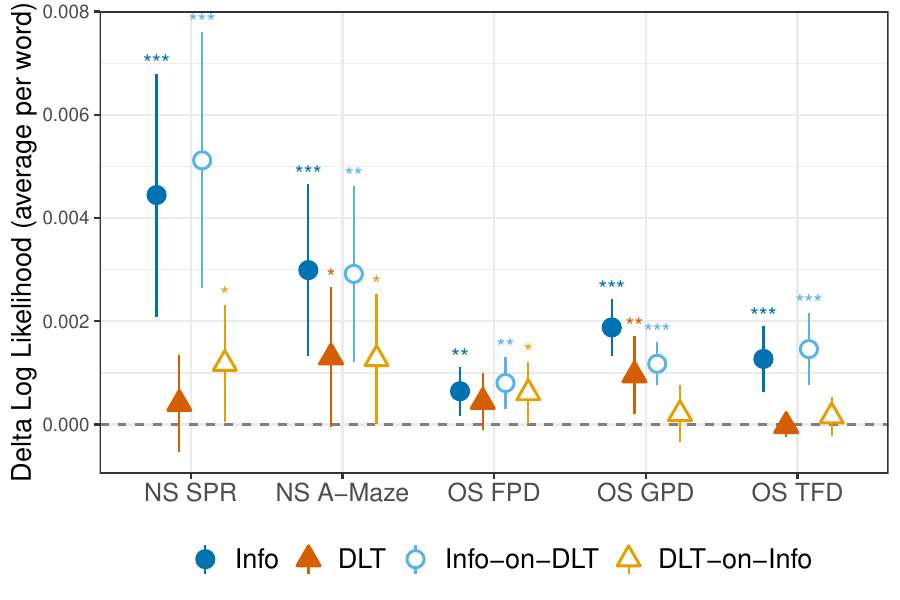}
        \caption{Predictive power of storage measures quantified by per-word \dll. Results are shown for four model comparisons. Points and error bars represent means and 95\% confidence intervals, respectively.
        Stars indicate FDR-adjusted significance levels (${}^{\star\star\star}$ $q<.001$, ${}^{\star\star}$ $q<.01$, ${}^\star$ $q<.05$, n.s. $q\geq 0.05$).}
        \label{fig:dll}
    \end{subfigure}
    \hfill
    \begin{subfigure}[t]{0.48\textwidth}
        \centering
        \includegraphics[width=\linewidth]{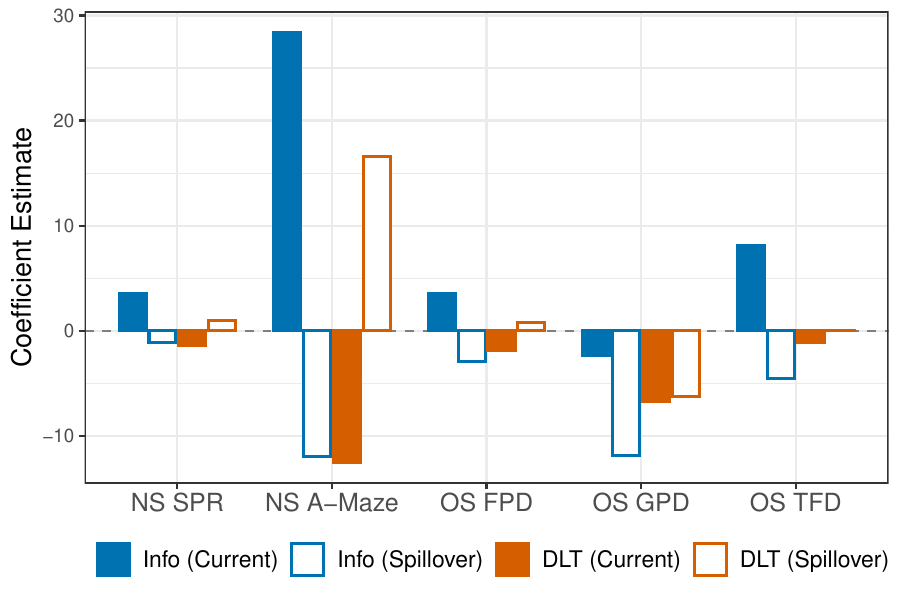}
        \caption{Mean regression coefficients for information storage and DLT storage at the current and spillover regions, averaged across 10 CV folds in the \Info and \DLT conditions, respectively. As the predictors were $z$-scored, coefficients represent relative effect sizes.}
        \label{fig:coef}
    \end{subfigure}
    
    \caption{
        Results of the naturalistic reading-time analysis on Natural Stories (NS) and OneStop (OS).
        Abbreviations: SPR = self-paced reading; FPD = first-pass duration; GPD = go-past duration; TFD = total fixation duration.
    }
    \label{fig:analysis_results}
\end{figure*}

\Cref{fig:dll} illustrates the predictive contributions of each storage measure.
Individually, information storage (\Info) significantly improved model fit in all five datasets, while DLT storage (\DLT) is significant in Natural Stories A-Maze and OneStop GPD.
The above-and-beyond analysis (\DLTInfo and \InfoDLT conditions) further reveals that the two measures capture largely independent aspects of reading-time variance.
Adding information storage to a model that already contains DLT storage always yields significant improvements, whereas adding DLT storage to a model that includes information storage results in significant improvements in three out of five tests.
Crucially, the magnitude of improvement for one measure remains largely stable regardless of whether the other is included in the model (compare \Info vs.\ \DLTInfo, and \DLT vs.\ \InfoDLT).
This consistent pattern suggests that while our information-theoretic measure significantly predicts reading times, it and the grammar-based DLT are complementary, each accounting for distinct sources of reading-time variance.
This is not necessarily surprising, given that LLM predictions are driven by far more than just structural information~\citep{hu2026can,mcgee-etal-2026}, and conversely, humans may rely more on structured information than LLMs~\citep{kajikawa-isono-ur}.


\Cref{fig:coef} shows the mean regression coefficients for information and DLT storage across folds of data for our linear regression models in the \Info and \DLT conditions, respectively.
For information storage, coefficients in the current region are positive for all datasets except for OneStop GPD, supporting the hypothesis that higher information storage increases processing difficulty.
For OneStop GPD, the coefficient is negative. This divergence between GPD and TFD suggests a specific reading strategy: high storage cost may prompt readers to move their gaze rightward more quickly to resolve uncertainty, subsequently leading to the regressions reflected in the increased total fixation duration.
As for DLT storage, the coefficients in the current region are consistently negative.
Recent studies on Japanese, an SOV language, suggest that readers exhibit differing processing strategies in response to high DLT storage cost, resulting in variable directions of the effect~\citep{isono-etal-ur,isono-kajikawa-2026}.
Against this backdrop, it remains an important question for future research to explore whether our negative effect reflects a uniform speed-up strategy among readers on average or stems from language-specific structural factors.

\section{Discussion}

In this study, we introduced \emph{information storage cost}, an information-theoretic measure of working memory storage quantifying expectations about future input.
Unlike traditional storage cost metrics that rely on specific syntactic theories and discrete counting, our measure is continuous, grammar-independent, and estimable from neural language models.
Our analyses demonstrated the validity of this approach: information storage successfully recovers the processing difficulty of center embeddings and object relative clauses solely from distributional statistics, correlates with grammar-based DLT storage, and predicts reading times in naturalistic text over and above baseline models.
These results suggest that the cognitive bottleneck of storage can be effectively operationalized as the maintenance of (half-pointwise) mutual information between the context and the future.

The current theory is noteworthy in the context of \emph{lossy-context surprisal}~\citep{futrell-etal-2020-lossy}.
Our measure operationalizes storage cost as the sum of predictive potentials under the assumption of perfect context maintenance.
It quantifies the informational load required to sustain the high-fidelity representations that underlie prediction errors in traditional surprisal theory \citep{hale-2001,levy-2008}.
However, human sentence processing is fundamentally constrained by memory loss and noise~\citep{lewis-vasishth-2005,levy-etal-2009}.
Just as lossy-context surprisal~\citep{futrell-etal-2020-lossy} generalized traditional surprisal by integrating memory constraints through noisy context representations, our framework provides a foundation for extending storage cost to account for such imperfect maintenance.
Thus, investigating \emph{lossy-context} storage remains a crucial next step to more accurately capture how the cognitive system manages information under finite resources.
Ultimately, while prior research on the \emph{memory-surprisal tradeoff}~\citep{hahn-etal-2021,hahn-etal-2022-morpheme} has primarily focused on corpus-level averages, our proposed metric provides the formal machinery to investigate these information-theoretic dynamics at the level of incremental, word-by-word sentence comprehension.

The observed alignment between our information-theoretic measure and grammar-based DLT storage is theoretically linked to the Head-Dependent Mutual Information (HDMI) hypothesis proposed by \citet{futrell-etal-2019}.
The HDMI hypothesis posits that syntactic dependencies exist primarily between word pairs exhibiting high PMI.
Indeed, \citet{futrell-etal-2020-lossy} provided empirical evidence across 54 languages showing that words in a syntactic dependency share higher PMI on average than those without such a relation.
This theoretical framework explains why our information-theoretic metric, despite being derived purely from distributional statistics, correlates with storage costs grounded in dependency grammar.

However, our analysis of naturalistic reading times revealed that this alignment is somewhat orthogonal to their predictive power: information storage and DLT storage capture partially distinct sources of reading-time variance.
This finding implies that the information-theoretic approach is not merely a continuous generalization that subsumes discrete grammar-based metrics.
Crucially, the fact that DLT storage remains a robust predictor even after controlling for information storage highlights the importance of structure-based metrics.
Given the fundamental bottleneck of working memory~\citep{christiansen-chater-2016}, it is not surprising that sentence comprehension involves abstract structural knowledge alongside statistical patterns.
To cope with the rapid loss of linguistic input, the cognitive system likely relies on symbolic structural representations to chunk or compress information efficiently---a function that probabilistic prediction based on statistical patterns alone may not fully fulfill.
Thus, statistical prediction and structural processing appear to operate as distinct, complementary mechanisms in overcoming the cognitive constraints of language comprehension.

\section{Limitations}\label{sec:limitation}
This study has several limitations, which are important to acknowledge.
We defined information storage at the word level, assuming that comprehenders maintain specific lexical predictions.
However, cognitive resource-rationality in sentence processing suggests that memory representations are likely compressed to optimize the tradeoff between precision and capacity~\citep{hahn-etal-2022-resource,xu-futrell-2026}.
Future work should explore the optimal granularity of compressed representations for modeling storage cost, rather than relying on raw tokens.

When estimating values of information storage using BERT, we assumed conditional independence between these tokens.
While this is a strong and unrealistic assumption for natural language, overcoming it within our current formulation presents a severe computational hurdle.
Since our definition of predictive potential requires estimating the joint probability of the future sequence conditioned on a gapped context (see \cref{eq:predpotential}), evaluating this quantity with autoregressive language models is theoretically possible but computationally prohibitive, as it necessitates marginalizing over the entire vocabulary at the omitted position \targetid at every incremental step.
As a radically alternative direction, rather than engineering a workaround for this computational dilemma, one could reconsider the underlying theoretical formulation itself.
Instead of aggregating the word-by-word predictive potentials, storage cost at $\curid$ could be reformulated at a holistic level as the \emph{half-pointwise predictive information} (mutual information) between the observed context and the future sequence:
\begin{align}
    &\mi{\Context=\context}{\Future} \notag \\
    &= \condE{\future}{\context}{\pmi(\context; \future)} \notag \\
    &= \KL{\condprob{\Future}{\context}}{\prob{\Future}}.
\end{align}
Crucially, this shift would eliminate the conditional independence assumption entirely, albeit at the expense of a distinct computational bottleneck: the need to approximate the unconditional marginal distribution of the future sequence, \prob{\Future}.

Finally, the interplay between storage and prediction likely varies across languages, particularly in head-final (SOV) languages like Japanese, where pre-verbal memory demands are structurally higher~\citep{nakatani-gibson-2010,isono-etal-ur,isono-kajikawa-2026}.
Investigating these compression strategies and cross-linguistic variations offers a promising avenue for refining information-theoretic models of working memory.



\section*{Acknowledgments}
We are grateful to Lin Ai and Taiga Someya for their valuable comments on earlier versions of this work.
This work is supported by JSPS KAKENHI Grant Number JP25K22996.

\bibliography{ref}

\clearpage
\onecolumn
\appendix
\crefalias{section}{appendix}
\crefalias{subsection}{appendix}

\section{Predictive Potential as KL Divergence}\label{app:kl}

Derivations establishing the equivalence of predictive potential to a KL divergence are provided.

Starting from the definition of contextualized PMI:
\begin{align}
    \condpmi{\wtarget}{\future}{\contextwotarget}
    &\coloneqq \log \frac{\condprob{\wtarget, \future}{\contextwotarget}}{\condprob{\wtarget}{\contextwotarget}\cdot \condprob{\future}{\contextwotarget}} \notag\\
    &= \log \frac{\condprob{\future}{\contextwotarget, \wtarget}\cdot \cancel{\condprob{\wtarget}{\contextwotarget}}}{\cancel{\condprob{\wtarget}{\contextwotarget}}\cdot \condprob{\future}{\contextwotarget}} \notag\\
    &= \log \frac{\condprob{\future}{\context}}{\condprob{\future}{\contextwotarget}}.
\end{align}

Taking the expectation over \future conditioned on \context:
\begin{align}
    \chmi{\wtarget}{\Future}
    &\coloneqq \condE{\future}{\context}{\condpmi{\wtarget}{\future}{\contextwotarget}} \notag\\
    &= \sum_{\future\in\Sigma^*} \condprob{\future}{\context} \log \frac{\condprob{\future}{\context}}{\condprob{\future}{\contextwotarget}} \notag\\
    &= \KL{\condprob{\Future}{\context}}{\condprob{\Future}{\contextwotarget}} \geq 0.
\end{align}
Non-negativity follows from the non-negativity of KL divergence.

\section{Monotonic Non-Increase in Expectation}
\label{app:mono}

We show that the predictive potential, the contextualized half-pointwise mutual information, is monotonically non-increasing in expectation over the next word.
In other words, the following relation holds:
\begin{align}
    \condE{\wcur}{\context}{\chmi{\wtarget}{\FutureNext}}\leq \chmi{\wtarget}{\Future}.
\end{align}

\paragraph{Chain rule for KL divergence.}
We first introduce the chain rule for KL divergence.
For any joint distributions $P(X, Y)$ and $Q(X, Y)$, applying the definition of KL divergence and the factorization $P(X, Y) = P(X)P(Y \mid X)$, we have:
\begin{align}
    &\KL{P(X, Y)}{Q(X, Y)} \notag\\
    &= \sum_{x, y} P(x, y) \log \frac{P(x, y)}{Q(x, y)} \notag\\
    &= \sum_{x, y} P(x, y) \log \frac{P(x) P(y \mid x)}{Q(x) Q(y \mid x)} \notag\\
    &= \sum_{x, y} P(x, y) \log \frac{P(x)}{Q(x)} + \sum_{x, y} P(x, y) \log \frac{P(y \mid x)}{Q(y \mid x)} \notag\\
    &= \KL{P(X)}{Q(X)} + \sum_{x, y} P(x) P(y \mid x) \log \frac{P(y \mid x)}{Q(y \mid x)} \notag\\
    &= \KL{P(X)}{Q(X)} + \sum_{x} P(x) \sum_{y} P(y \mid x) \log \frac{P(y \mid x)}{Q(y \mid x)} \notag\\
    &= \KL{P(X)}{Q(X)} + \E{X}{\KL{P(Y \mid X)}{Q(Y \mid X)}}.
    \label{eq:kl_chain}
\end{align}
This states that the divergence between joint distributions equals the divergence between marginals plus the expected divergence between conditionals.

The predictive potential of \wtarget at position \curid is:
\begin{align}
    \chmi{\wtarget}{\Future} = \KL{\condprob{\Future}{\context}}{\condprob{\Future}{\contextwotarget}}.
\end{align}

Similarly, at position \nextid:
\begin{align}
    \chmi{\wtarget}{\FutureNext} = \KL{\condprob{\Futurenext}{\contextnext}}{\condprob{\Futurenext}{\contextwotargetnext}}.
\end{align}

We decompose \chmi{\wtarget}{\Future} using the chain rule for KL divergence~\eqref{eq:kl_chain}.
Note that $\Future = (W_k, \FutureNext)$.
Applying the chain rule:
\begin{align}
    \chmi{\wtarget}{\Future}
    &= \KL{\condprob{\Wcur,\Futurenext}{\context}}{\condprob{\Wcur,\Futurenext}{\contextwotarget}} \notag\\
    &= \KL{\condprob{\Wcur}{\context}}{\condprob{\Wcur}{\contextwotarget}} \notag\\
    &+ \condE{\wcur}{\context}{\KL{\condprob{\Futurenext}{\context,\wcur}}{\condprob{\Futurenext}{\contextwotarget,\wcur}}}.
\end{align}

Conditioning on $W_{\curid} = \wcur$, we have $\contextnext = (\context, \wcur)$ and $\contextwotargetnext = (\contextwotarget, \wcur)$.
Thus, the KL divergence inside the expectation equals \chmi{\wtarget}{\FutureNext} evaluated at context \contextnext.

Since KL divergence is always non-negative, we obtain:
\begin{align}
    \condE{\wcur}{\context}{\chmi{\wtarget}{\FutureNext}}\leq \chmi{\wtarget}{\Future}.
\end{align}

This shows that, in expectation over the word $\Wcur=\wcur$, the predictive potential of \wtarget decreases (or stays the same) as the sentence proceeds.
Of course, for a specific observed word \wcur, the value \chmi{\wtarget}{\FutureNext} may exceed \chmi{\wtarget}{\Future}, but \chmi{\wtarget}{\FutureNext} does not exceed \chmi{\wtarget}{\Future} on average.

To verify this theoretical result empirically, we approximate the expectation via a Monte Carlo estimate using the UD\_English-GUM.
Specifically, we compute the sample mean of predictive potentials as a function of distance $d = k - i$.
For each distance $d$ up to 30, we collect all \chmi{\wtarget}{\Future}.

\begin{figure}[ht]
    \centering
    \includegraphics[width=0.42\linewidth]{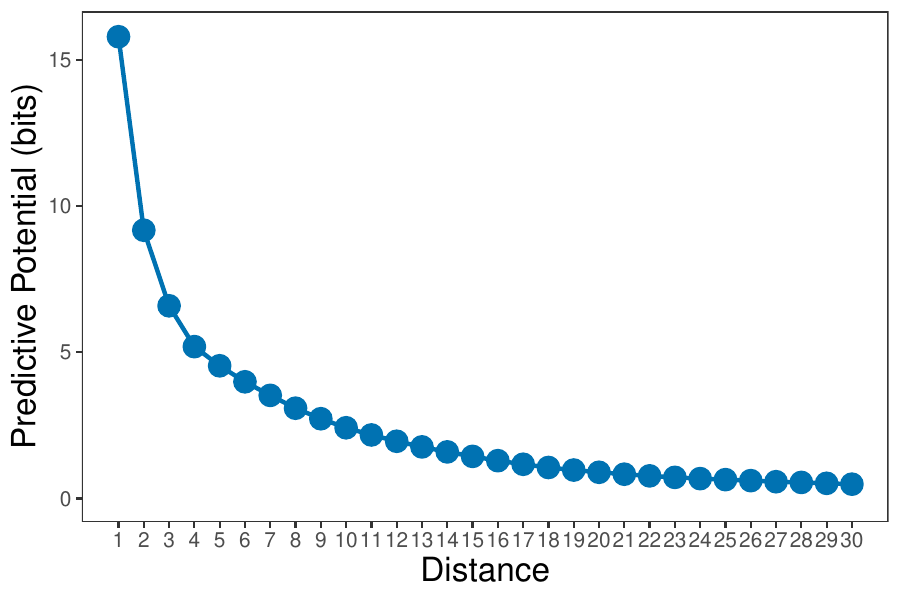}
    \caption{Mean predictive potential \chmi{\wtarget}{\Future} as a function of distance $d = k - i$ (up to a maximum of 30), computed over the UD\_English-GUM. The monotonic decrease is consistent with the theoretical result that predictive potential is non-increasing on average.}
    \label{fig:distance_decay}
\end{figure}

\Cref{fig:distance_decay} shows that the mean predictive potential decreases monotonically with distance, consistent with the theoretical prediction.
This decay pattern reflects the general property that words become less predictive of future materials as the distance gets longer.

\section{Generated Stimulus Materials}\label{app:sents}

The complete set of 30 items used for the illustrative analyses in \cref{sec:illustration} is listed. Each item consists of two syntactically distinct but lexically identical sentences.

\paragraph{Center-embedding vs.\ right-branching structures}
The following items contrast double center-embedded structures (a) with their corresponding right-branching variants (b).

\begin{enumerate}[label={(\arabic*)}, nosep]
  \small
  \item a. The actor who the doctor who the reporter watched visited called the dancer \\ b. The reporter watched the doctor who visited the actor who called the dancer
  \item a. The lawyer who the dancer who the doctor attacked stopped praised the writer \\ b. The doctor attacked the dancer who stopped the lawyer who praised the writer
  \item a. The judge who the pilot who the senator watched visited attacked the reporter \\ b. The senator watched the pilot who visited the judge who attacked the reporter
  \item a. The nurse who the thief who the reporter avoided praised visited the guard \\ b. The reporter avoided the thief who praised the nurse who visited the guard
  \item a. The singer who the guard who the pilot called questioned admired the student \\ b. The pilot called the guard who questioned the singer who admired the student
  \item a. The reporter who the artist who the singer ignored called helped the pilot \\ b. The singer ignored the artist who called the reporter who helped the pilot
  \item a. The banker who the detective who the doctor helped attacked warned the president \\ b. The doctor helped the detective who attacked the banker who warned the president
  \item a. The neighbor who the thief who the teacher stopped admired praised the senator \\ b. The teacher stopped the thief who admired the neighbor who praised the senator
  \item a. The doctor who the chef who the president avoided watched called the guard \\ b. The president avoided the chef who watched the doctor who called the guard
  \item a. The thief who the neighbor who the judge met attacked praised the banker \\ b. The judge met the neighbor who attacked the thief who praised the banker
  \item a. The writer who the student who the soldier attacked visited watched the president \\ b. The soldier attacked the student who visited the writer who watched the president
  \item a. The chef who the teacher who the baker ignored helped watched the actor \\ b. The baker ignored the teacher who helped the chef who watched the actor
  \item a. The neighbor who the banker who the writer watched avoided praised the teacher \\ b. The writer watched the banker who avoided the neighbor who praised the teacher
  \item a. The president who the thief who the actor visited praised stopped the artist \\ b. The actor visited the thief who praised the president who stopped the artist
  \item a. The artist who the baker who the chef praised avoided trusted the teacher \\ b. The chef praised the baker who avoided the artist who trusted the teacher
  \item a. The guard who the student who the writer trusted noticed watched the detective \\ b. The writer trusted the student who noticed the guard who watched the detective
  \item a. The senator who the student who the detective visited attacked called the chef \\ b. The detective visited the student who attacked the senator who called the chef
  \item a. The judge who the singer who the detective warned admired avoided the banker \\ b. The detective warned the singer who admired the judge who avoided the banker
  \item a. The actor who the baker who the lawyer praised visited ignored the teacher \\ b. The lawyer praised the baker who visited the actor who ignored the teacher
  \item a. The guard who the teacher who the dancer trusted questioned warned the judge \\ b. The dancer trusted the teacher who questioned the guard who warned the judge
  \item a. The neighbor who the student who the lawyer noticed attacked admired the nurse \\ b. The lawyer noticed the student who attacked the neighbor who admired the nurse
  \item a. The senator who the doctor who the lawyer avoided noticed ignored the actor \\ b. The lawyer avoided the doctor who noticed the senator who ignored the actor
  \item a. The pilot who the thief who the president admired questioned warned the chef \\ b. The president admired the thief who questioned the pilot who warned the chef
  \item a. The nurse who the teacher who the guard attacked praised avoided the reporter \\ b. The guard attacked the teacher who praised the nurse who avoided the reporter
  \item a. The writer who the guard who the teacher called attacked helped the actor \\ b. The teacher called the guard who attacked the writer who helped the actor
  \item a. The pilot who the artist who the baker called trusted praised the reporter \\ b. The baker called the artist who trusted the pilot who praised the reporter
  \item a. The nurse who the artist who the doctor avoided watched called the actor \\ b. The doctor avoided the artist who watched the nurse who called the actor
  \item a. The nurse who the thief who the banker ignored noticed helped the lawyer \\ b. The banker ignored the thief who noticed the nurse who helped the lawyer
  \item a. The guard who the nurse who the reporter met admired helped the thief \\ b. The reporter met the nurse who admired the guard who helped the thief
  \item a. The doctor who the neighbor who the soldier questioned visited met the student \\ b. The soldier questioned the neighbor who visited the doctor who met the student
\end{enumerate}

\paragraph{Subject vs.\ object relatives}
The following items contrast subject relative clauses (a) with object relative clauses (b).
\begin{enumerate}[label={(\arabic*)}, nosep]
  \small
  \item a. The actor who called the doctor praised the reporter \\ b. The actor who the doctor called praised the reporter
  \item a. The student who avoided the lawyer attacked the dancer \\ b. The student who the lawyer avoided attacked the dancer
  \item a. The dancer who warned the guard questioned the president \\ b. The dancer who the guard warned questioned the president
  \item a. The senator who trusted the reporter visited the president \\ b. The senator who the reporter trusted visited the president
  \item a. The nurse who visited the thief stopped the reporter \\ b. The nurse who the thief visited stopped the reporter
  \item a. The singer who visited the actor warned the guard \\ b. The singer who the actor visited warned the guard
  \item a. The baker who watched the judge noticed the teacher \\ b. The baker who the judge watched noticed the teacher
  \item a. The reporter who helped the artist warned the singer \\ b. The reporter who the artist helped warned the singer
  \item a. The teacher who helped the lawyer noticed the banker \\ b. The teacher who the lawyer helped noticed the banker
  \item a. The doctor who helped the president attacked the chef \\ b. The doctor who the president helped attacked the chef
  \item a. The neighbor who met the thief noticed the teacher \\ b. The neighbor who the thief met noticed the teacher
  \item a. The dancer who warned the baker attacked the guard \\ b. The dancer who the baker warned attacked the guard
  \item a. The president who avoided the guard watched the soldier \\ b. The president who the guard avoided watched the soldier
  \item a. The thief who praised the neighbor visited the judge \\ b. The thief who the neighbor praised visited the judge
  \item a. The president who noticed the senator visited the writer \\ b. The president who the senator noticed visited the writer
  \item a. The soldier who attacked the president watched the student \\ b. The soldier who the president attacked watched the student
  \item a. The chef who watched the teacher avoided the baker \\ b. The chef who the teacher watched avoided the baker
  \item a. The neighbor who called the artist avoided the banker \\ b. The neighbor who the artist called avoided the banker
  \item a. The singer who questioned the writer attacked the actor \\ b. The singer who the writer questioned attacked the actor
  \item a. The actor who visited the artist praised the guard \\ b. The actor who the artist visited praised the guard
  \item a. The artist who avoided the baker called the chef \\ b. The artist who the baker avoided called the chef
  \item a. The singer who helped the guard avoided the student \\ b. The singer who the guard helped avoided the student
  \item a. The senator who called the student warned the detective \\ b. The senator who the student called warned the detective
  \item a. The president who praised the banker trusted the judge \\ b. The president who the banker praised trusted the judge
  \item a. The detective who warned the banker admired the actor \\ b. The detective who the banker warned admired the actor
  \item a. The actor who ignored the baker called the lawyer \\ b. The actor who the baker ignored called the lawyer
  \item a. The student who called the dancer stopped the guard \\ b. The student who the dancer called stopped the guard
  \item a. The dancer who questioned the judge trusted the pilot \\ b. The dancer who the judge questioned trusted the pilot
  \item a. The chef who stopped the neighbor ignored the student \\ b. The chef who the neighbor stopped ignored the student
  \item a. The driver who attacked the president watched the senator \\ b. The driver who the president attacked watched the senator
\end{enumerate}

\section{Correlations between Predictors for Naturalistic Reading Analysis}
\label{app:cor}

\begin{figure}[ht]
    \centering
    \begin{subfigure}[b]{0.53\textwidth}
        \centering
        \includegraphics[width=\linewidth]{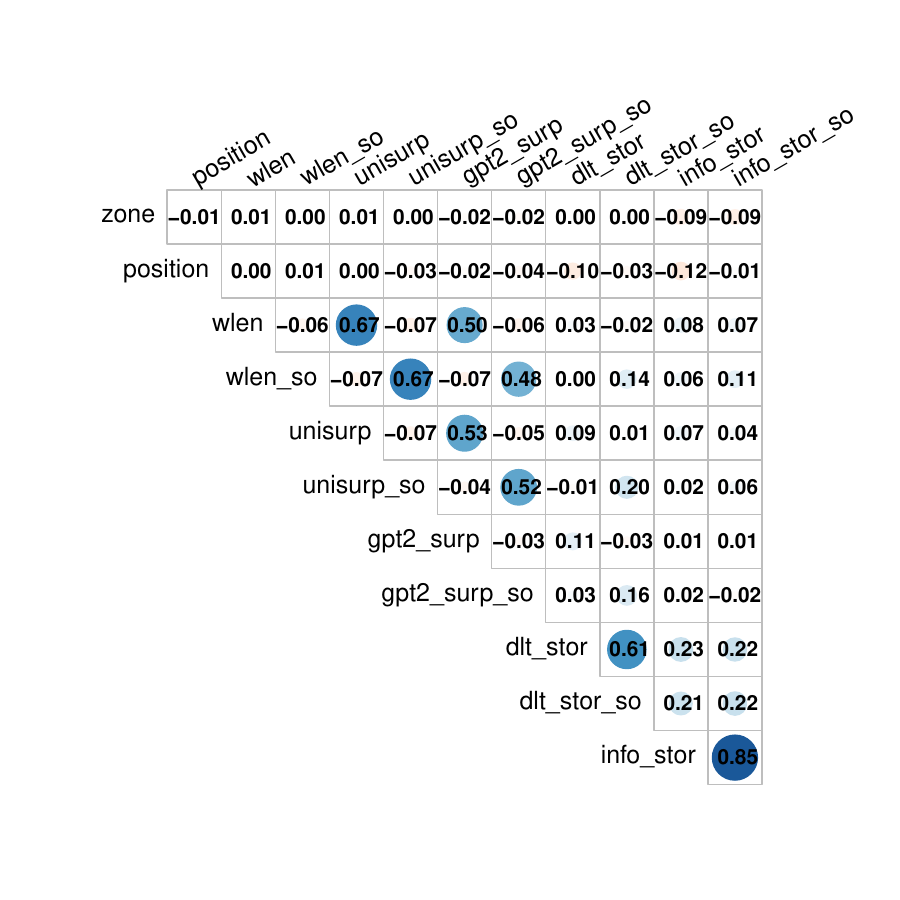}
        \caption{Natural Stories}
        \label{fig:ns_cor}
    \end{subfigure}
    \hfill
    \begin{subfigure}[b]{0.53\textwidth}
        \centering
        \includegraphics[width=\linewidth]{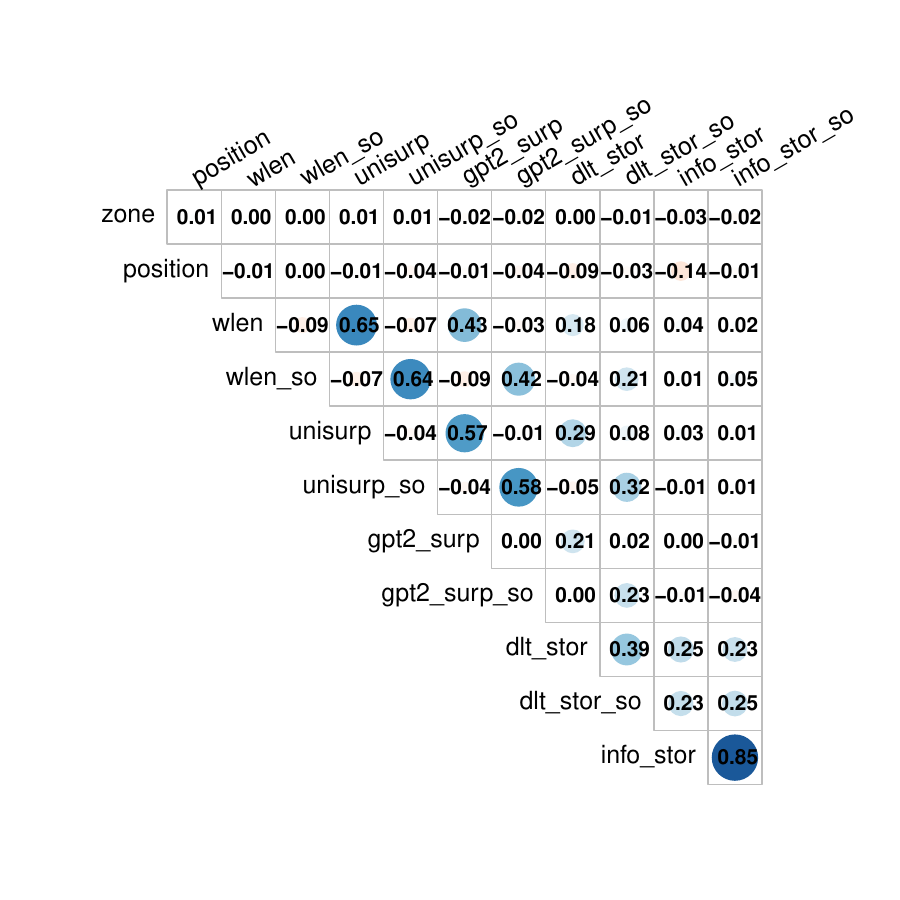}
        \caption{OneStop}
        \label{fig:os_cor}
    \end{subfigure}
    \caption{Pearson correlation matrices of predictors in the naturalistic reading-time datasets.}
    \label{fig:large_cor}
\end{figure}

We present the Pearson correlation matrices for the predictors used in the analysis of the Natural Stories and OneStop datasets in \cref{fig:large_cor}.
The abbreviations for the predictors are defined as follows:
\begin{itemize}[noitemsep]
    \item \texttt{zone}: Word position index in the document.
    \item \texttt{position}: Word position index within the sentence.
    \item \texttt{wlen}: Word length in characters.
    \item \texttt{unisurp}: Unigram surprisal.
    \item \texttt{gpt2\_surp}: Surprisal estimated by GPT-2 small.
    \item \texttt{dlt\_stor}: Storage cost based on Dependency Locality Theory.
    \item \texttt{info\_stor}: Information-theoretic storage cost (proposed).
    \item \texttt{\_so}: The suffix appended to variables to denote their corresponding spillover terms.
\end{itemize}

\section{Comparison with other BERT-family models}\label{app:bertlargeroberta}

While the main text reports the information storage estimates using the BERT-base model (\texttt{bert-base-uncased}), here we present a comparison with estimates derived from other models.
Specifically, we report the information storage estimates by BERT-Large (\texttt{bert-large-uncased}) and RoBERTa (\texttt{roberta-base};~\citealp{liu2019roberta}).
The architectural specifications of each model are summarized in \cref{tab:bertcaps}.

\begin{table}[ht]
    \centering
    \begin{tabular}{lcccc}
        \toprule
        Model & Layers & Heads & Hidden Size & Parameters \\
        \midrule
        BERT-base (\texttt{bert-base-uncased})    & 12 & 12 & 768   & 110M \\
        BERT-Large  (\texttt{bert-large-uncased}) & 24 & 16 & 1,024 & 336M \\
        RoBERTa (\texttt{roberta-base})           & 12 & 12 & 768   & 125M \\
        \bottomrule
    \end{tabular}
    \caption{Architectural details of the models. Layers, Heads, and Hidden Size refer to the number of layers, the number of attention heads per layer, and embedding/hidden size, respectively.}
    \label{tab:bertcaps}
\end{table}

\subsection{Correlation to DLT storage}
\begin{figure}[ht]
    \centering
    \begin{subfigure}[b]{0.4\textwidth}
        \centering
        \includegraphics[width=\linewidth]{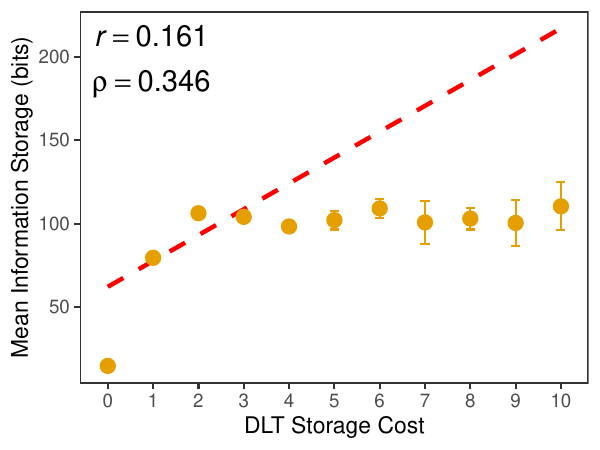}
        \caption{BERT-Large}
        \label{fig:bertlarge_cor}
    \end{subfigure}
    \hfill
    \begin{subfigure}[b]{0.4\textwidth}
        \centering
        \includegraphics[width=\linewidth]{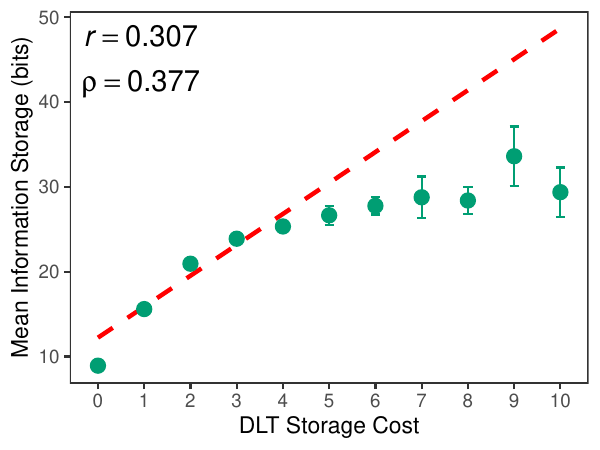}
        \caption{RoBERTa}
        \label{fig:roberta_cor}
    \end{subfigure}
    \caption{Mean information-theoretic storage cost as a function of DLT storage cost in the UD\_English-GUM corpus. Points represent the mean value for each DLT bin with 95\% confidence intervals. For this visualization, bins with fewer than 100 observations were excluded due to data sparsity. The red dashed line represents the linear regression fitted to the raw data.}
    \label{fig:infoscorr}
\end{figure}

Following the methodology described in~\cref{sec:correlation}, we computed information storage on the UD\_English-GUM corpus using BERT-Large and RoBERTa, and investigated its correlation with DLT storage cost.
The results are illustrated in \cref{fig:infoscorr}. 

For RoBERTa, although the correlation coefficients decreased slightly compared to the BERT-base results reported in \cref{sec:correlation}, the overall trend remained largely consistent.
In the case of BERT-Large, while the Pearson correlation coefficient was lower than those of the other models, the Spearman correlation remained above $0.3$.
This suggests that the relationship between information storage and DLT storage is not strictly linear, despite exhibiting a clear monotonically increasing trend.

\subsection{Reading-time analysis}
\begin{figure}[ht]
    \centering
    \begin{subfigure}[t]{0.49\textwidth}
        \centering
        \includegraphics[width=\linewidth]{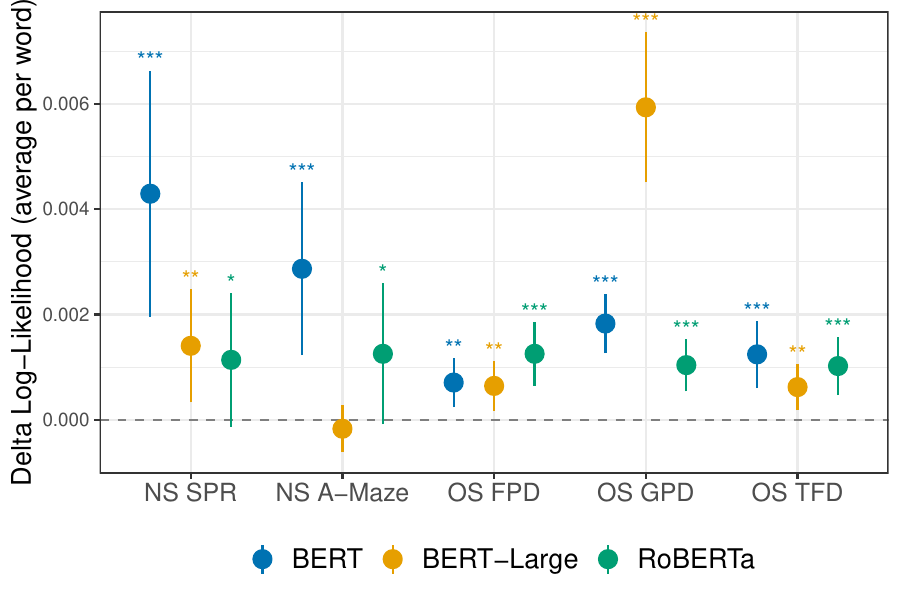}
        \caption{Predictive power of information storage measures quantified by per-word \dll. Information storage is estimated from three different BERT models: BERT (\texttt{bert-base-uncased}), BERT-Large (\texttt{bert-large-uncased}), and RoBERTa (\texttt{roberta-base}). Points and error bars represent means and 95\% confidence intervals, respectively.
        Stars indicate FDR-adjusted significance levels (${}^{\star\star\star}$ $q<.001$, ${}^{\star\star}$ $q<.01$, ${}^\star$ $q<.05$, n.s. $q\geq 0.05$).}
        \label{fig:dllinfos}
    \end{subfigure}
    \hfill
    \begin{subfigure}[t]{0.49\textwidth}
        \centering
        \includegraphics[width=\linewidth]{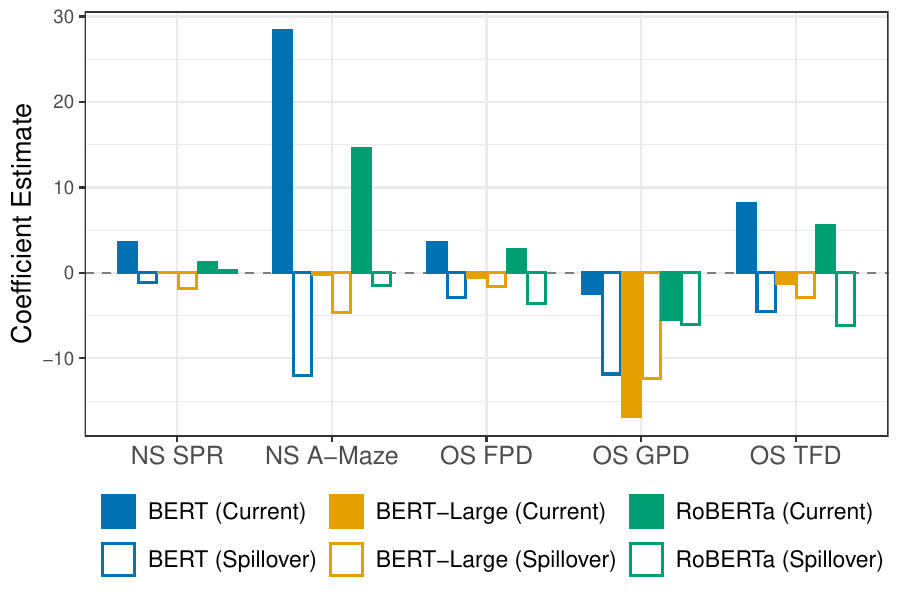}
        \caption{Mean regression coefficients for information storage at the current and spillover regions, averaged across 10 CV folds. As the predictors were $z$-scored, coefficients represent relative effect sizes.}
        \label{fig:coefinfos}
    \end{subfigure}
    \caption{Results of the naturalistic reading-time analysis on Natural Stories (NS) and OneStop (OS).
        Abbreviations: SPR = self-paced reading; FPD = first-pass duration; GPD = go-past duration; TFD = total fixation duration.}
    \label{fig:infosrt}
\end{figure}

We replicated the reading-time analysis from \cref{sec:rtanalysis} using the information storage estimates derived from the different BERT models.
Specifically, we investigated whether information storage exhibits significant predictive power for reading times and examined the resulting coefficients.
The results are presented in \cref{fig:infosrt}.

The results for RoBERTa largely aligned with those for BERT-base.
Although the magnitude of the effects varied, the \dll values were all significantly greater than zero, indicating significant predictive power for reading time.
Furthermore, the directions of the coefficients were entirely consistent, with the sole exception of the spillover coefficient in the Natural Stories SPR dataset.

For BERT-Large, \dll indicated significant predictive power for all datasets except Natural Stories A-Maze; however, the coefficients were predominantly negative.
Given that DLT storage also exhibited a negative tend (see \cref{fig:coef}), the interpretation of these coefficients warrants further examination in future work.

Overall, BERT-base and RoBERTa demonstrated highly similar behavior.
In contrast, while BERT-Large often patterned with BERT-base, their behavior was not always perfectly aligned.
This discrepancy can likely be attributed, at least in part, to the differences in parameter counts (see \cref{tab:bertcaps}).
This observation may be related to empirical findings on causal language models, which demonstrate that surprisal estimates from models with approximately 125M parameters provide the best fit for reading time in English, whereas larger models yield poorer fits~\citep{oh-schuler-2023}.

\end{document}